\documentclass[format=acmsmall, review=true, screen=false]{acmart}
\setcitestyle{numbers,sort&compress}
\usepackage{natbib}
\usepackage{booktabs} 
\usepackage[utf8]{inputenc}
\usepackage[noend]{algpseudocode}

\makeatletter
\def\BState{\State\hskip-\ALG@thistlm}
\makeatother

\DeclareMathAlphabet{\mathcal}{OMS}{cmsy}{m}{n}

\usepackage[ruled]{algorithm2e} 

\acmJournal{TIST}
\acmVolume{9}
\acmNumber{4}
\acmArticle{1}
\acmYear{2018}
\acmMonth{12}

\setcopyright{acmcopyright}

\acmDOI{0000001.0000001}

\begin{document}
\title{A Zero-Shot Learning application in Deep Drawing process using Hyper-Process Model}
\lhead{Zero-Shot Learning in Deep Drawing}

\author{João Reis}
\email{jpcreis@fe.up.pt}
\author{Gil Gonçalves}
\email{gil@fe.up.pt}
\affiliation{%
  \institution{SYSTEC, Research Center for Systems and Technologies}
  \department{Faculty of Engineering, University of Porto}
  \city{Porto}
  \country{Portugal}
}

\begin{abstract}

One of the consequences of passing from mass production to mass customization paradigm in the nowadays industrialized world is the need to increase flexibility and responsiveness of manufacturing companies. The high-mix / low-volume production forces constant accommodations of unknown product variants, which ultimately leads to high periods of machine calibration. The difficulty related with machine calibration is that experience is required together with a set of experiments to meet the final product quality. Unfortunately, all possible combinations of machine parameters is so high that is difficult to build empirical knowledge. Due to this fact, normally trial and error approaches are taken making one-of-a-kind products not viable. Therefore, a Zero-Shot Learning (ZSL) based approach called hyper-process model (HPM) to learn the relation among multiple tasks is used as a way to shorten the calibration phase. Assuming each product variant is a task to solve, first, a shape analysis on data to learn common modes of deformation between tasks is made, and secondly, a mapping between these modes and task descriptions is performed. Ultimately, the present work has two main contributions: 1) Formulation of an industrial problem into a ZSL setting where new process models can be generated for process optimization and 2) the definition of a regression problem in the domain of ZSL. For that purpose, a 2-d deep drawing simulated process was used based on data collected from the Abaqus simulator, where a significant number of process models were collected to test the effectiveness of the approach. The obtained results show that is possible to learn new tasks without any available data (both labeled and unlabeled) by leveraging information about already existing tasks, allowing to speed up the calibration phase and make a quicker integration of new products into manufacturing systems.

\end{abstract}

%
%

\ccsdesc[500]{Computing methodologies~Transfer learning}
\ccsdesc[500]{Applied computing~Command and control}

%
%

\keywords{Zero-Shot Learning, hyper-modeling, smart manufacturing}

\maketitle

\section{Introduction} \label{intro_rw}

After mass production, which was the norm in the late 20\textsuperscript{th} century, and mass customization in the early 21st century, we are now in a period where personalized production is considered a commodity. From cars to sneakers, from furniture to clothes, every buyer wants to select and adjust to its needs or tastes every single detail of the product he/she is buying. Small lots of customer-oriented products, with short delivery times and high-mix/low-volume production are among the current challenges industry is facing. Industry 4.0 and related technologies are enablers for new manufacturing systems capable of dealing with these challenges of high-mix/low-volume production. This production follows the observed paradigm shift from mass production to mass customization where any product can be specific and one-of-a-kind to fulfill customers’ requirements. In order to achieve that, knowledge about equipment and processes is peremptory to increase the flexibility in nowadays manufacturing systems.

This way, one of the cornerstones of personalized production is the area of machine learning, where high-volumes of data can be processed and interpreted for wiser and quicker decision making. This is mostly relevant in the context of production systems that regularly yield new product variants where no experience exists, but lead time should be maintained. These situations force companies to be more flexible and drastically responsive in yielding and building a whole new set of products. Normally, the operation of a certain machine is guided by a set of process parameters that influence process quality and dictate the final result of a certain product. Therefore, the correct process parameters need to be chosen to yield the correct process quality subject to a set of process conditions. Hence, there is an implicit relation on how machine parameters influence the final quality of the product. This way, a good understanding of how these dynamics work is peremptory for process automation. We can first define a task as a function that maps equipment process parameters into process quality for a specific product variant. These are often named process models and they are used not only to predict the product quality based on certain parameters, but also to find the best process parameters according to some given process quality \cite{reis2018optimization, pollak2016models}. This means that customers can specify quality requirements for their products, and process models can be used to automatically find the best parameters for a specific machine that meets those requirements. To build these process models, a significant amount of data is required where a large set of parameters need to be tested, experiments performed and data collected in form of a dataset in order to effectively train a process model using machine learning techniques.

Although it seems that process models can solve the problem of high-mix / low-volume production, this is only part of the solution. The main limitation for these process models is that once trained, they are only applicable for a specific product variant and not extendable to new unseen product variants, even if using the same machine. This means that if a new product variant needs to be yielded, a whole new dataset is required to train a new process model. This is an impractical situation because the cost per experiment is very high and a significant amount of experiments is required, leading to long periods of data collection. Therefore, new and innovative methods are necessary to alleviate such a limitation and relate data that was already collected together with new sources of information.

The current paper proposes to solve such a process model limitation by exploring an area of machine learning called Zero-Shot Learning (ZSL) where the relations among multiple tasks are learned and extrapolations to new unseen tasks can be made. The ZSL problem is framed into a broader area of machine learning called Transfer Learning, where the goal is to assist on the learning process for a future problem of interest where the amount of data is scarce. Particularly for the ZSL problem, it is assumed that no data is available from target tasks during the training phase, which makes the problem much harder to solve. Hence, this very same principle can be applicable to manufacturing systems, and in particular to process optimization, where the relations between process models can be used to foresee the dynamics of future process models of interest. For instance, if there are process models A, B and C that correspond to product variants yielded in the same machine (in this context, different product variants might represent different metal sheet materials for welding, different thicknesses, different shapes, etc.), ZSL can be used to gain some knowledge about an unseen product variant D by predicting the corresponding process model. Through this new process model the best parameters can be estimated based on a set of target process quality specified by the customer.

The present work has two main contributions regarding the state of the art: 1) formulating a ZSL problem for an industrial process optimization scenario, and 2) how ZSL can be formulated considering regression scenarios. Regarding the first contribution, the main intention is to start fostering the use of ZSL in industry as a way to develop new intelligent systems that could significantly improve the nowadays systems' performance. By formulating the problem, we hope to help other researchers to do the same kind formulation for their industrial application scenarios more easily. The second contribution is more related with the use of ZSL regardless of the application, since the problem in hands is of regression. This way, we hope to complement the area of ZSL to support both regression and classification problems, and ultimately apply these to all kind of real world problems.

This paper is organized in 5 more Sections. Section 2 presents the related work about the ZSL and Section 3 introduces of ZSL algorithm for regression problems named Hyper-Process Model. Section 4 explains the deep drawing process as well as the scenario and experimental setup used in the proposed approach. Section 5 presents all the results achieved in the present work and finally Section 6 closes the paper with a discussion and main conclusions.

\section{Related Work}

As previously described, the main purpose of Transfer Learning (TL) is to use past experience and knowledge to accelerate the process of learning a new task \cite{pan2010survey, weiss2016survey}. The tasks used to accelerate the learning of a future task are called the sources tasks, while the future task is named target task. Hence, the idea is to simply learn the relation about a set of source tasks and together with some indications about the target task, a better learning process can be taken. The area of TL is mainly composed by three major sub areas namely Inductive, Transductive and Unsupervised TL. Inductive TL is related with problems where there are available labeled data from source tasks, together with a limited amount of labeled data on the target task. Transductive TL is similar to the previous one, but normally only unlabeled data is available on the target task and concepts such as Domain Adaptation \cite{daume2006domain}, Covariate Shift \cite{shimodaira2000improving} and Sample Selection Bias \cite{zadrozny2004learning} are used. Unsupervised TL is related with a set of problems when only unlabeled data in both source and target tasks is available, and self-taught learning is normally applied \cite{raina2007self}.

In this work, we are particularly interested in the case where no data at all is available on the target task, but labeled data exists in the source tasks. This kind of problem is called Zero-Shot Learning (ZSL) where a prediction of the correct value or labeled is achieved using zero data from the future problem of interest. The area ZSL has been dominated in the past decade by classification problems, where a certain input should be correctly classified, even if the corresponding class was not used during training. However, most of the process models used in manufacturing systems are related with regression meaning that metrics of quality assessment are in form of continuous variables. This presents a gap in state of the art methods where no algorithms exist specifically for regression problems. Hence, the aim of the present section is to first describe the difference between classification and regression for ZSL problems, and secondly present the latest ZSL works and applicability where the great majority is related with image classification.

To that intent, most of the ZSL techniques take advantage on the difference between input images, which is something usual where two different objects are displayed. Assuming inputs for a certain class / task as $X_i \in \mathcal{X}$ for class $i$, we can say that these techniques assume $P(X_i) \neq P(X_j)$ where marginal distributions among classes are not the same. This means that the difference between the images can be learned to separate both from different classes, and vice-versa. Contrary to this, we should state that ZSL in regression problems the inputs for different tasks could be the same and the responses might be different according to their specific task. Most of the works first try to map the input into a latent space, which normally is a task descriptor (e.g. a semantic descriptor), that can be generally expressed as $\mathcal{G} : X^n \rightarrow F^k$, where $X \in \mathbb{R}^n$ are inputs and $F \in \mathbb{R}^k$ are the task descriptors. In order to successfully learn the differences between tasks or classes, there should exist some difference between the task inputs like cubes and spheres, or cats and houses. Therefore, the assumption of $P(X_i) \neq P(X_j)$ is implicit in the context of image classification, which might not hold true for regression. Hence, this draws the first difference between how ZSL works for classification and regression, where it is not assumed that the marginal distribution of inputs from different tasks are different. This is the scenario explored in the present paper, where we assume $P(X_i) = P(X_j)$.

Additionally, another key difference between ZSL for classification and regression that should be pointed out is how the algorithms operate once trained. The ultimate goal of ZSL in classification is to provide a new unseen image and correctly predict the label from a class not used in the learning process. Opposite to this idea, for the regression setting, the main goal is to build a whole new predictive function suitable for the new unseen task, where multiple inputs can be fed as a regular regressor. Therefore, for each source task, a regressor needs to be previously learned and together with the task description, a new predictor can be derived for a target task.

%
%

Up to our knowledge, there are a hand-full of works that handle the scenario of equal marginal distributions from inputs and are capable to derive new predictors. One of these works is presented by \citet{larochelle2008zero} where the authors present two different approaches to the ZSL problem: 1) input space view and 2) model space view. The first approach uses a concatenation of the input $x$ and the task / class description $d(z)$ for a given task $z$, and by using a supervised learning algorithm train a model $f^*(.)$ to predict $y_t^z$. The second approach is more model-driven, and is defined by $f_z(x) = g_{d(z)}(x)$. By defining a joint distribution $p(x,d(z))$ one can then set $g_{d(z)}(x) = p(x|d(z))$ and learn a probabilistic model that estimates the input $x$ belonging to class $d(z)$. However, a different way to achieve model space view is also presented. This uses the model parameters $\theta$ to train a model that maps class descriptions into model parameters, allowing to predict some target parameters according to new unseen conditions. The authors have applied these approaches to character recognition, handwritten characters and molecular compound application scenarios, where the benefits of models space view are depicted comparing with input space view.

Additionally, the same principle was applied to solve a concrete problem in the area of manufacturing systems named hyper-model (HM) \cite{pollak2016models}. In this work, the same principle as the model space view is applied, where model parameters are learned together with process conditions (task descriptions) so when new conditions arrive, the corresponding model parameters could be predicted. Despite these two works might be applicable to regression problems, they both suffer from the same limitation. Since the relation among model parameters and task description / process condition should be learned, all the source model parameters should have the same representation. This means that the same type of machine learning technique needs to be used among all source tasks, representing a great limitation if two different tasks have different dynamics and complexity, and one technique might suit better one task than other, and vice-versa. Hence, one of the works that propose to overcome such a limitation is introduced by \citet{reis2017meta}. In their work, an intermediate shape representation of data is used as a way to standardize the models and make them comparable, despite of the technique used. However, the authors did not frame the work in a ZSL setting, and the properties studied in the scenario are limited to a small number of process models, constraining the true evaluation of the proposed approach. In the present work, a new formulation of this approach is presented and applied to a more significant and complete industrial scenario. To that intent, data collected from the Abaqus simulator related with a 2-d deep drawing process was used, where a significant number of process models were trained to assess the impact of the algorithms' performance in generating new process models. This way, the study in the present work assesses the performance of the algorithm contemplating a varying number of source process models to generate new ones, and also the selection of source models that maximized the overall system performance.

%
%
Regarding the application of ZSL to classification, there are already a significant amount of important works. The work of \citet{palatucci2009zero} paved the way for a more formal definition and theory of zero-shot learning. In their work a two-stage approach is presented where the same concept as task description is used as before but now called semantic feature space. For their approach, the main idea is to map brain images into semantic feature space and then label these features according to available classes. The authors call the whole solution the semantic output code (SOC) classifier. The main reason to separate the learning into two stages and avoid training directly a function is to predict class labels that are not present in the training phase. Therefore, the goal is to train the first phase (brain images to semantic feature space) with a set of inputs that map into certain class labels, and then train the second stage (semantic feature space to labels) with a larger spectrum of class labels. 

%
%
A similar two-stage approach called cross-model transfer (CMT) was proposed by \citet{socher2013zero} where the main idea is to train a model that is able to map image features into a word vector space, and then have a second model that is trained to classify these word vectors into the correct label classes. Again, it is assumed that more classes are present in the second stage rather than in the first. In the first stage, based on the work of \citet{coates2011importance}, the authors have extracted a set of unsupervised image features from raw image pixels in order to map these into a semantic space (word vector) \cite{huang2012improving}. As for the second stage, the authors want to first assess if the presented image is from seen or unseen classes, so then labels can be chosen based on likelihood.

%
%

Another interesting work worth referring that is also related with this two-stage approach was first introduced by \citet{lampert2009learning} and then further extended by \citet{lampert2014attribute}, where two different techniques were presented: 1) direct attribute prediction (DAP); and 2) indirect attribute prediction (IAP). For the DAP technique, a probabilistic model was used to estimate the probability of binary-value attributes given a certain image, so unseen images at test phase could also have an estimate into this attribute space. The predictions from image to unseen class were then made using maximum a posteriori (MAP). As for the IAP technique, instead of a two-stage approach, an additional stage was used. First a mapping between image and training classes is performed, as a regular multiclass classifier, estimating the probability for each training class. Then, a mapping between training classes and attributes is made resulting in a model that maps images in attributes.

%
%
The work of \citet{qiao2016joint} presents an algorithm that was greatly inspired by the DAP algorithm, where the authors consider a chain of dependent attributes and the joint probability of each attribute for a specific class is calculated, contrary to DAP which calculates the marginal probability. However, due to high amount of attributes it is difficult to calculate these joint probabilities, so first a clustering algorithm is applied to organize attributes into sets.

%
%

First introduced by \citet{akata2013label} and then extended and generalized by the same group \cite{akata2016label}, the attribute label embedding (ALE) is presented as an alternative that outperforms some of the DAP method limitations \citet{lampert2014attribute}, namely: 1) a two-stage learning approach for ZSL problem, by 2) assuming the attributes on AwA are independent among themselves and 3) is not extendable to other sources of side information. For ALE implementation, an already existing algorithm called web-scale annotation by image embedding (WSABIE) proposed by \citet{weston2010large} was used as a baseline. For the optimization process, the authors use stochastic gradient descent (SGD) as a convex-function is not guaranteed. Moreover, the authors explore other kinds of embeddings such the hierarchical label embedding (HLE) first proposed by \citet{tsochantaridis2005large} or the word2vec label embedding (WLE) proposed by \citet{frome2013devise}, resulting in these 4 algorithms were considered: DAP, ALE, HLE, WLE.

%
%

The work presented by \citet{akata2015evaluation} proposes a new approach called structured joint embedding (SJE). The difference between SJE and the previously presented ALE algorithm is mainly on the optimization function, where the authors preferred the unregularized structured SVM. The authors also present a additional approach based on multiple output embeddings. The algorithm learns the best transformation $W$ for a specific output embedding, and according to the given input embedding the best class is selected based on a confidence in each of the embeddings. As a certain output embedding can benefit more some classes than others, this approach uses multiple output embeddings and learns the best according to the provided input embedding.

%
%
The approach presented by \citet{xian2016latent} is called latent embeddings (LatEm), and is a direct extension of the SJE where a nonlinear piece-wise compatibility function is explored, opposed to the linear one used in SJE. This nonlinear compatibility is explored by learning a collection of linear models, where each linear model maximizes the compatibility among image-class embedding pairs.

Some other interesting works were also proposed for the image classification problem in ZSL and worth mentioning, such as the deep visual-semantic embedding model \cite{frome2013devise}, the joint latent similarity embedding (JLSE) \cite{zhang2016zero}, the convex combination of semantic embeddings (CONSE) \cite{norouzi2013zero}, the semantic similarity embedding (SSE) \cite{zhang2015zero}, the embarrassingly simple approach to zero-shot learning (ESZSL) \cite{romera2015embarrassingly}, the synthesized classifiers (SYNC) \cite{changpinyo2016synthesized}, the semantic autoencoder for zero-shot learning (SAE) \cite{kodirov2017semantic}, the simple exponential family framework (GFZSL) \cite{verma2017simple}, the zero-shot classification with discriminative semantic representation learning (DSRL) \cite{ye2017zero}, the feature generating networks (FGN) \cite{xian2018feature} and the gaze embeddings (GE) for zero-shot image classification \cite{karessli2017gaze}. For a comprehensive survey of ZSL methods for image classification please refer to the work presented by \citet{xian2018zero}.

%
%

%
%

%
%

%
%

%
%

%
%

%
%

%
%

%
%

%
%

%
%
One of the most interesting applications of ZSL outside image classification domain is related with object identification using haptic devices presented by \citet{abderrahmane2018haptic}. In their work, the authors use the DAP algorithm, and different variations of such, to recognize a set of objects by grasping those with a robotic hand with tactile fingertips. The main idea behind the ZSL setting is to be able to correctly recognize an object that the system was not trained for. Hence, from cutaneous and kinesthetic information of the robotic hand, the system should correctly say what object is holding, e.g. a plastic bottle, lamp or cup of tea, without any prior information about this specific object.

%
%
In line with the previous works, \citet{isele2016using} present an approach where the model parameters of a policy based approach in a reinforcement learning (RL) setting is predicted based on a set of defined task descriptors. This work makes use of the same principle as by \citet{larochelle2008zero} and \citet{pollak2016models}, but the methods used to achieve it are different. The main goal of the present work is to jointly learn a sparse encoding of both model parameters from a policy and task descriptors in a latent representation. To this joint learning the author call \textit{coupled dictionary learning} and to the whole algorithm task descriptors for lifelong learning (TaDeLL). The rational behind such algorithm is that similar task descriptions have similar policies, so information can be learned from these two different spaces with an adaptation to the policy gradient (PG), introduced by \citet{sutton2000policy}. The authors perform a set of tests in three different simulated environments: 1) spring mass damper (SM); cart pole (CP); and 3) bicycle (BK).


\section{Hyper-Process Model} \label{chap_hp}

This section will focus on the explanation of the approach used for ZSL with application in the deep drawing process. As previously explained, the presented algorithm is a refinement and better realization of a prior work \cite{reis2017meta} where in this paper we name it as hyper-process model (HPM), and is composed by two different methods. First we will introduce the hyper-model concept \citep{pollak2016models} for process models in manufacturing applications, and secondly present the statistical shape model (SSM) \citep{cootes1995active} for image segmentation. Ultimately, the HPM can be viewed as an extension to the hyper-model it self, and hence its name.

\subsection{Proposed Approach} \label{proposed_approach}

Algorithm \ref{zsl_algorithm} presents all the steps required to implement the HPM for different contexts of application. Therefore, the first thing to notice is that the algorithm itself is divided into two different parts, one making use of the statistical shape model (SSM) \citep{cootes1995active} and the other using the hyper-model \cite{pollak2016models}. The SSM is a widely used technique for image segmentation that analyzes the geometrical properties of a set of given shapes or objects by creating deformable models using statistical information. As a mathematical transform to be applied to these set of shapes, the most common techniques are principal component analysis, approximated principal geodesic analysis, hierarchical regional PCA \citep{mesejo2016survey} and singular value decomposition, where non-affine modes of deformation are calculated. The same way this method assumes that there exist specific shape variations and these can be quantified forming a deformable model, is the same way that we assumed that these variations also exist in different tasks, and a deformable model for a set of tasks can be derived.

The hyper-model \cite{pollak2016models} is a concept where a model of models is built and is applied to industrial scenarios. Complementarily, the authors introduce the notion of \textit{condition} that are fixed quantities that govern a certain industrial process, like thickness in metal sheets for welding processes or deep drawing. This concept of condition is what defines each task in the context of ZSL, where different conditions mean different tasks. Assuming that a model is a set of base functions that transforms a certain input into an output, a model has always associated a condition that quantitatively describes the task to learn. Based on this, the main idea is to build a hyper-model to generate models for a whole continuum of conditions, aiming at mapping model coefficients from the base functions into a set of conditions. This way, by providing a set of new conditions, it is possible to derive a new set of coefficients and build a new model for prediction in the context of those conditions only. As one might have realized by now, this approach is independent from being a regression or classification problem. As far as the coefficients of base functions and conditions are available, the hyper-model can be applied to both settings. 

Based on this, the algorithm starts to introduce all the parameters necessary for its execution. As described, all the trained models are required along with the corresponding task descriptions (which are the conditions in the hyper-model approach). Moreover, the target task description is required in order to generate the new model. Additionally, one should also specify the number of landmarks to use for each shape, together with two more vectors that define the minimum and maximum values for the input features space. These minimum and maximum vectors are required so one could generate the input values to sample from the trained models. Since we are assuming $P(X_s) = P(X_t)$, only a vector is required and the minimum and maximum values will be used in all source tasks to generate shapes. Finally, we assume to have $m$ trained models to deal with.

For this algorithm, the SSM first comes into place because the hyper-model is dependent on the common representation of models to be trained. Hence, the first step (line 1) is to generate the input values $X$ according to the minimum, maximum and number of intended landmarks per shape. Since we assume that no information can be drawn between the different inputs from the various models (as stated by $P(X_i) = P(X_j)$), the same input values are used for all the models. Therefore, the shapes $S_i$ are built only considering the values from the output feature space, as presented in line 3, where $i$ is a specific model.

\begin{algorithm}[h!]
\SetAlgoNoLine
\caption{Hyper-Process Modeling} \label{zsl_algorithm}
\KwData{HPM($F,\varsigma,\varsigma',n,min,max$) - $F$ is a set of source models, $\varsigma$ is a set of task descriptions associated with each source model, $\varsigma'$ is the target task description to be used for model generation, $n$ is the number of data points per shape, $min$ and $max$ are vectors of size $r$ (assuming $	 X_i \in \mathbb{R}^r$) with minimum and maximum values for the input features, correspondingly. Finally, $m$ in the number of source models.}

\KwResult{Generated model $f'$}

\textit{Statistical Shape Model:} \\
\Indp
\nl Define the input to sample from existing models: $X \gets GenerateInput(min,max,n)$\;
\nl \For {$i = 1 \rightarrow m$}{
    \nl Get shape: $S_i = f_i(X)$
}
\nl Get the mean shape: $\bar{S} = \frac{1}{N} \sum_{i=1}^{N} S_i$\;
\nl Get eigenvectors from PCA decomposition: $\phi \gets PCA(S)$\;
\nl Get deformable parameters from PDM: $\boldsymbol{b} = \boldsymbol{\phi}^T ( S - \bar{S})$\;
\Indm
\textit{Hyper-Model:} \\
\Indp
\nl Train the hyper-model: $h : b \rightarrow \varsigma$\;
\nl Get the deformable parameter for new shape: $b' = h^{-1}(\varsigma')$\;
\nl Get new shape: $S' = \bar{ \boldsymbol{S}} +  \boldsymbol{\phi}  b'$\;
\nl Train a model for the new task. $f' : X \rightarrow S'$\;

\nl \Return $f'$\;

\end{algorithm}

The next step is to calculate the mean shape from all the generated shapes (line 4). In order to get all the eigenvectors to build the deformable model, a decomposition needs to be performed on all the generated shapes and PCA is applied (line 5). One should emphasize again that each shape is a vector of $kn$ elements, where $k$ is the number of features and $n$ is the number of landmarks to use. Therefore, PCA is performed on a $m \times kn$ matrix $S$ composed by all the shapes from source models, where these shapes are stacked in rows. Finally, the last step for the SSM is to derive all the deformable parameters for all the models (line 6). These are the parameters required to generate back the initial shape based solely on the deformable model. In order to get a good shape reconstruction, the number of components chosen when performing PCA is critical, being a trade-off between reconstruction and complexity. On one hand, if few components are chosen, the greater the reconstruction error will be but less dimensions are required, and thus, less complex the problem is. On the other hand, if all the components are chosen, the reconstruction error will be minimum, but the complexity of the problem is far too great to deal with. In these situations, a good rule of thumb is to use the number of components (ordered by decreasing order of model variance) that attend for a cumulative sum of variance of at least 95\%.

After building the deformable model, together with all the deformable parameters, the hyper-model is ready to be trained. For this case, and as presented by \citet{pollak2016models}, one should train a hyper-model using any machine learning technique that seems suitable for the problem, by mapping deformable parameters into task descriptors (conditions in the context of the hyper-model). One might think at this stage that would be more suitable to map task descriptors into deformable parameters instead, because we can use the trained model to predict the parameters based on new descriptors. However, in most of the cases the dimension of the deformable parameters are greater than descriptors, so the modeling needs to be made according to line 7. Only in the cases where 1) the dimension of parameters is the same or lower than descriptors or 2) multiple models are trained as a hyper-model and each one of those models has only an output variable, and each single output is different from the ones in other models, the hyper-model can be trained as follows $h : \varsigma \rightarrow b$. The implication of building a hyper-model that maps deformable parameters into descriptors is visible in line 8, where the technique used needs to be invertible in order to get the new deformable parameters according to the specified new descriptors. As an alternative, the level set where the model surface intercepts with the hyper-plane for the intended target descriptors can be calculated, as performed in the work of \citet{pollak2011retrieval}, or formulate a minimization problem where the distance between the predicted and target descriptors should be minimized \cite{reis2018laser, reis2018optimization}. Once the deformable parameters are obtained from the hyper-model according to the target descriptors, the next step is to generate a new shape as presented in line 9. The last step is to train a model to map the initially generated input values into the generated shape, which corresponds to the output values for that specific task description.

\section{Zero-Shot Learning in Deep Drawing process} \label{deep_drawing_section}

The main purpose of the present section is to make use of the HPM algorithm to generate a new model upon new task descriptors and therefore, be able to predict the final quality of a future product variant of interest. To this intent, a simulated scenario with a high number of models will be used in the context of the deep drawing process. Hence, a description of the process will be made, followed by the process modeling phase where multiple source models are trained using ANNs, and finally the generation of unseen models using solely new task descriptors will be presented.  

\subsection{Deep Drawing process}

The deep drawing process is a widely-used manufacturing procedure that aims at forming a blank metal sheet into a cup or box-like shapes using a press machine. This way is possible to form any flat metal sheet into the desired shape, which occurs under a combination of tensile and compressive conditions leading ultimately to the manufacture of light weight and low density products. For that, a great knowledge of the process is required since a high number of parameters can be optimized, from blank-holder force / pressure, punch radius, die radius, material properties just as elasticity, to coefficient of friction used \cite{dwivedi2017study}. Pots and pans for cooking, containers, sinks, automobile parts, such as panels and gas tanks, are just some examples of the products manufactured by sheet metal deep drawing \citep{deepdrawing}.

\begin{figure}[ht]
\vskip 0.2in
\begin{center}
\centerline{\includegraphics[width=5.5in]{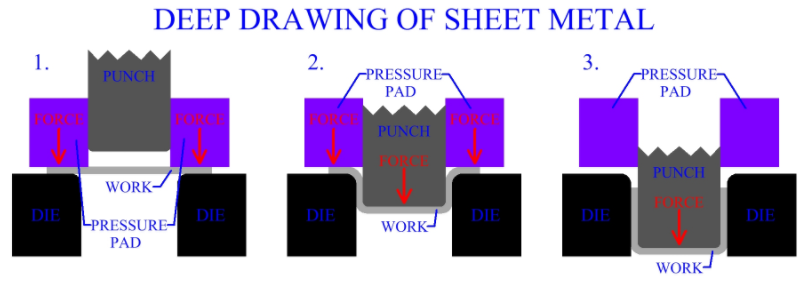}}
\caption{Exemplification of a Deep Drawing process in 3 main steps: 1) First position of the piece; 2) Punch force to form the flat metal sheet; 3) Final positioning and end of the process \citep{deepdrawing}}
\label{deep_drawing_process}
\end{center}
\vskip -0.2in
\end{figure}

As can be seen in Figure \ref{deep_drawing_process}, this process is mainly composed by 4 main components: blank (light gray) - metal-sheet to form; punch (dark gray) - tool that forces the forming of the metal-sheet and molds it; die (black) - base plate that supports the metal-sheet during forming; blank-holder (purple) - applies pressure in the metal-sheet extremities against the die. Normally, the process is composed by three main steps. The first step is the initial setup of the machine where the metal sheet is flat and the force is applied in its extremities to hold the sheet. This force is made from a pressure pad against the die. The second step is mainly composed by continuously applying a punch force downwards to mold the metal sheet into the desired final shape. Finally, the last step is when the punch stops applying pressure and the final shape is achieved.

\subsection{Scenario Description}

The scenario explored in the present work is related with the generation of process models that map a set of process parameter values into a product quality measures. One pivotal aspect that will be studied is how HPM deals with a varying number of source models. This way, a considerable number of process models need to be trained together with the process conditions that explicitly describe a certain process. To that intent, a simulator called Abaqus was used to prepare a suitable set of datasets that describe the deep drawing process with a large combination of process parameters and conditions.

Abaqus is a simulator for finite element analysis (FEA) and computer aided engineering (CAD / CAM). For this case, the "Complete Abaqus Environment" or Abaqus/CAE version 6.14 was used where a simulation of mechanical component models can be executed and an analysis of its performance can be done. For the present work, a 2-d simulation was used with a metal sheet and three more components that allow to shape this sheet. These components are namely the punch, pressure pad and die according to the deep drawing process described previously.

In order to train a process model, the process parameters and corresponding process quality should be well defined. Therefore, there are two process parameters that can be changed yielding different quality in the final product. These parameters are the Blank Holder Force (BHF) - force applied by the pressure pad into the metal sheet against the die -  and Friction (F) - amount of friction in the die that allows the metal sheet to move more smoothly or roughly. A low value for BHF leads to wrinkling in the metal-sheet, while a high value for BHF leads to fracture where the blank can break. Additionally, the lack of friction between the blank sheet and both die and blank holder can lead to earring on top of the formed piece. This way, it is very important to optimize these parameters during a calibration phase whenever a new variant needs to be yielded. As for the process quality, we have assessed two different measures. One of them is the Maximum Stress (S) observed during the whole process and the other one is the Distance (D) of the metal sheet from the initial to the final positions. This means that the process models can predict the maximum stress and metal sheet distance according to the parameters used, for specific process conditions, which in the context of ZSL are the task descriptors.

Complementary, we have defined three task descriptors that change the relation between process parameters and quality. These are the Metal Sheet Thickness (T), the Initial Stress (IS) and the Saturation (S). As for the IS and S, these are parameters used in the plasticity model during Abaqus simulation of the process, where the isotropic hardening law of Hockett-Sherby was used \citep{hockett1975large}. Therefore, the IS is the parameter for initial value of the yield stress while S is the flow stress saturation value. Assuming that a machine learning algorithm is used to map process parameters into quality, different values of task descriptors imply a different distribution of data, and hence a new predictor needs to be trained. From Table \ref{deep_process_conditions} all the combination of deep drawing task descriptors are depicted, where 18 processes will be used to test the proposed approach.

\begin{table}[h]
\centering
\caption{Process Conditions for the Deep-Drawing process simulation in Abaqus}
\label{deep_process_conditions}
\begin{tabular}{lccc}
\hline \\
           & Metal Sheet Thickness (T) & Initial Stress (IS) & Saturation (S) \\
\hline
Process 1  & 1.5                   & 100            & 130        \\
Process 2  & 1.5                   & 100            & 165        \\
Process 3  & 1.5                   & 100            & 200        \\
Process 4  & 1.5                   & 175            & 227.5      \\
Process 5  & 1.5                   & 175            & 288,75     \\
Process 6  & 1.5                   & 175            & 355        \\
Process 7  & 1.5                   & 250            & 325        \\
Process 8  & 1.5                   & 250            & 412.5      \\
Process 9  & 1.5                   & 250            & 500        \\
Process 10 & 2                     & 100            & 130        \\
Process 11 & 2                     & 100            & 165        \\
Process 12 & 2                     & 100            & 200        \\
Process 13 & 2                     & 175            & 227.5      \\
Process 14 & 2                     & 175            & 288.75     \\
Process 15 & 2                     & 175            & 355        \\
Process 16 & 2                     & 250            & 325        \\
Process 17 & 2                     & 250            & 412.5      \\
Process 18 & 2                     & 250            & 500       \\
\hline
\end{tabular}
\end{table}

In order to obtain the 18 models, the Abaqus Simulator was used with each of the task descriptor and a set of values for process parameters BHF and F. For each descriptor, 6 different values for BHF were used: 0.66, 0.1, 0.133, 0.167, 0.2 and 0.23; as for F, 9 different values were used: 0.12, 0.23, 0.34, 0.45, 0.56, 0.67, 0.78, 0.89 and 1. This means that a dataset of $6 \times 9 = 54$ datapoints per task description was created to train a process model. Taking in mind that each datapoint is a single simulation in Abaqus, there were required a total of $54 \times 18 = 972$ simulations to create the baseline for the current scenario. The values for BHF and F were chosen based on empirically evaluating the maximum of the process parameters where the maximum stress was so high that the end result was undesired deformations of the metal sheet or a fracture. All these 18 process conditions imply different relations between process parameters and quality. This way, for each set of task descriptions we need to model this unique relation in a process model.



As detailed in the beginning of the present section, the main idea of using a high number of different deep drawing processes is to assess how HPM deals with a varying number of source models and impacts the final performance of generated models. Our hypothesis states that with an increase of source models in HPM there will be a decrease of Mean Squared Error (MSE), and vice-versa. In this context, the MSE is the difference between the generated process model and the dataset ground truth. For ZSL, the intuition is that by increasing the number of source tasks, the interpolation among these tasks will become increasingly better and the overall error for the target predictor generation will decrease. This will allow to study the applicability of such an approach from problems with small number of models to more complex ones with a significantly higher number of source models available.

This way, and according to the proposed approach, the first step to apply the HPM is to train a set of process models using a machine learning algorithm from the available datasets and therefore build a deformable model based on SSM. This deformable model requires a set of shapes that will be generated from the trained source models. The next sub section will depict the machine learning algorithm used for modeling as well as the whole training procedure. 


\subsection{Process Modeling}

The present section aims at presenting the process of training a machine learning algorithm to map process parameters into process quality for the deep drawing scenario. As explained in the previous section, there are 18 processes that can be modeled into process models and generate the corresponding shapes for the SSM. The process parameters of Blank Holder Force (BHF) and Friction (F) are the independent variables (inputs) for the model, and process quality of Maximum Stress (S) and Distance (D) are the dependent variables (outputs). This way is possible to predict the quality of the process based solely on a set of possible machine parameter values. 

\begin{figure}[h]
\vskip 0.2in
\begin{center}
\centerline{\includegraphics[width=5.5in]{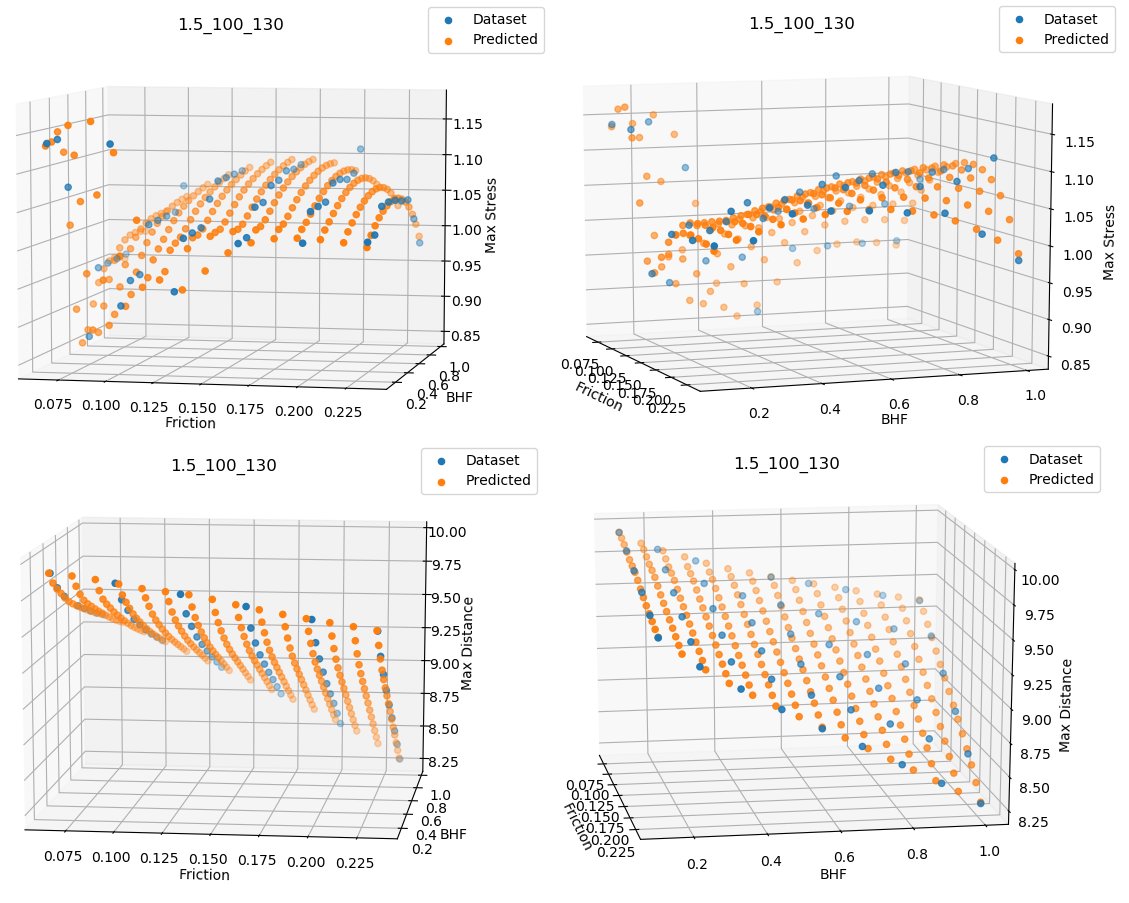}}
\caption{3D Scatter plot for the Process Parameters Blank Holder Force and Friction, and the Process Quality Maximum Stress (above) and Distance (below) for the following task descriptors: Metal Sheet Thickness 1.5, Initial Stress of 100 and Saturation of 130.}
\label{trained_ann_15_100_130_all}
\end{center}
\vskip -0.2in
\end{figure}

As for the chosen algorithm for the modeling process, feedforward multi-layer perceptron was used, also widely known as Artificial Neural Networks (ANN), with backpropagation for parameter optimization. For this particular case, a fixed topology was used with 3 hidden layers and 10 neurons each, along with the a learning rate starting at 1 with the Adaptive Subgradient Methods for weight optimization \cite{duchi2011adaptive} and decreased once two consecutive epochs fail to decrease the training loss by 1e-8, and 150,000 epochs were used for the training process. This optimization method was mainly used to avoid overfitting to data and ensure convergence for the parameter optimization. In this particular case, no k-fold cross-validation was performed due to the reduced number of datapoints and the complexity of the data to model. Instead, a visual inspection was performed, as depicted in Figure \ref{trained_ann_15_100_130_all}, in order to ensure that the model corresponds to a good generalization of the system dynamics. The MSE obtained from training all the 18 models is very low, being the highest MSE of 6.96e-4, meaning that the model could in fact fit the presented data and be a good generalization. In the presented Figure, the used task descriptors are T = 1.5, IS = 100 and S = 130, where the predicted values are depicted as orange dots and the ground truth values as blue.


\subsection{Process Model Generation} \label{process_generation_deep}

In order to generate process models based on unseen process conditions and existing process models, the proposed algorithm of HPM should be applied. In this case, unseen means that data for the corresponding task were not used during the training phase. As detailed in the previous sections, the main idea is to use a technique from computer graphics called SSM that based on a set of shapes, a mean shape and a set of deformable parameters are calculated to predict new related shapes. The great advantage of this method is that we can vary the deformable parameters within a certain range to create new shapes that are different, but related to those used in the SSM. 

In order to create the shapes for the SSM, we need to use the process models to sample a new dataset. As described previously, shapes have two assumption: 1) All have the same number of datapoints and 2) all datapoints from a shape should have a direct correspondence. This way, an input dataset (Friction and Blank Holder Force) was created in order to obtain different output shapes (Maximum Stress and Distance). As previously explained, there must be differences in shapes in order to calculate the corresponding deformable parameters. Hence, by fixing the input values for all process models, the main differences in responses can be assessed. In a practical way, this means that all 18 models correspond to 18 different product variants that are yielded in the same machine, and thus the machine parameters and its range are the same. This way, as only the output will be different for different product variants, a shape is defined only with the output of the process models.

As for the defined inputs of the process models, a dataset was built using a full factorial design (FFD) approach. This method allows to grasp the overall distribution of response variables according to a full variation of process parameters. The FFD is dependent on levels and factors. Factors are simply the number of process parameters that one wants to vary, and the levels are the number of equally distributed points for each process parameter to perform an experiment. Assuming a full combination of all the process parameter values, the number of datapoints is determined using $l^n$, where $l$ is the number of levels and $n$ is the number of factors. For this case, since the inputs are Friction and Blank Holder Force, the number of factors is 2 and we have defined a number of levels as 15 to have a large and complete representative shape. This means that the number of datapoints per shape is $15^2 = 225$, which is more than 4 times larger than the initial datasets. The orange dots on the scatter plot from Figure \ref{trained_ann_15_100_130_all} also represent the shape composed from 225 points.

Before moving forward, for this particular case we should clarify an important point about the generation of shapes. One might question why we need to train the process models at this stage to generate a set of shapes and not simply use the 54 datapoint datasets sampled from the Abacus simulator directly in the SSM as shapes. Despite being a reasonable thing to do in this very specific context, this is not usually the case at shop-floor. Normally, different people perform a different amount of experiments with different input values for machine parameters, which invalidates using these experiments as shapes due to 1) different amount of datapoints and 2) non-matching datapoints throughout the datasets. From this point of view, the trained process models can be seen as a transformation from the initial datasets into suitable shapes.


Regarding SSM, all the shapes have 225 datapoints sampled from the process models, where the statistical model and the corresponding deformable parameters $\boldsymbol{b}$ need to be calculated. AS already explained, one key aspect is the number of components from the PCA decomposition that needs to be performed in all shapes. Therefore, to determine the relevant number of these components / eigenvectors to use, a preliminary assessment was made. This assessment increases the number of components from 1 to 10 and calculates the MSE between all real and generated shapes. Ultimately, the ideal number of components for this scenario is 3 where the increase in the number of components did not decrease significantly the overall MSE.

\begin{figure}[h]
\vskip 0.2in
\begin{center}
\centerline{\includegraphics[width=4in]{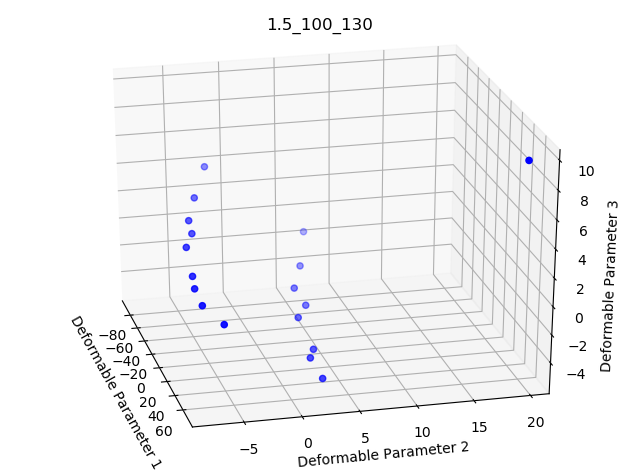}}
\caption{3D scatter plot for 3 deformable parameter regarding all 18 shapes from the deep drawing processes.}
\label{deformable_parameters}
\end{center}
\vskip -0.2in
\end{figure}

Since the main idea is to build an hyper-model for HPM based on the task descriptions and deformable parameters, in order to better understand the relation among the shapes in the deformable space, a 3-d scatter plot is built. Figure \ref{deformable_parameters} depicts all the 3 deformable parameters. It can be clearly seen that there are two groups of points referring to the two different existing thicknesses, being clear that this task descriptor is greatly correlated with deformable parameter 2. Additionally, there is an outlier in the top right corner, where it is visibly far from the two observed clusters because, in order to generate back the respective shape, the parameters should be considerably different. This means that the shape is too different from the mean calculated shape for the deformable model, and hence too different from the remaining ones. This shape refers to the task description where T = 2, IS = 100 and S = 130 and it was not considered for the next steps of the HPM approach and only 17 processes were used.

Based on the definition of the hyper-model \citep{pollak2016models}, a mapping of process model parameters into process conditions should be made using a machine learning algorithm. This means that for ZSL, in order to predict the deformable parameters for the SSM based on a new task description, an optimization problem needs to be formulated where the distance between predicted and target task description is minimized, finding the most suitable deformable parameters. However, since in this case we have 3 task descriptors and 3 deformable parameters, we could simply train a machine learning model inversely. Instead of mapping process parameters into task descriptions, we can map task descriptions directly in process parameters and avoid the minimization problem. Considering task descriptions as $\boldsymbol{\varsigma}_n$ and deformable parameters as $\lambda_n$, the new equation used for the hyper-model is as follows:

\begin{equation}\label{hyper_model_inv} \boldsymbol{\lambda} =  g_{\boldsymbol{\beta}} (\boldsymbol{\varsigma}) \end{equation}

As for the next step of the proposed approach, the hyper-model should be trained. For this case, three different statistical methods were used, namely Linear Regression Model \cite{lai1979strong}, Least Absolute Shrinkage and Selection Operator (also known as LASSO regression) \cite{tibshirani1996regression} and Ridge Regression \cite{hoerl1970ridge} - which is based on Tikhonov regularization and in machine learning mostly known as weight decay for ANNs. For the linear regression, a polynomial of degree 1, 2 and 3 were used, and for both LASSO and ridge only polynomials of degree 2 and 3 were used. The reason behind choosing these models lie in the existence of a small amount of datapoints (processes) to train.

For a better understanding of the tests to be performed, we should refer to the linear regression as LIN, degree 2 polynomial as POL2, degree 3 polynomial as POL3, LASSO using a degree 2 polynomial as LASSO2, LASSO using a degree 3 polynomial as LASSO3, and the same for Ridge, where RIDGE2 and RIDGE3 are the degree 2 and 3 polynomial, correspondingly. This leads to the test of 7 different approaches for the hyper-model in the HPM.


\section{Results}

Complementary to all the conditions aforementioned about all the regression methods to be used, 2 different tests will be performed in order to better understand the dynamics of deep drawing regarding different task descriptions. In order to test the hyper-model performance using a varying number of processes, one should define a distance measure between the processes to choose the closest processes to the one being generated and assess the improvement of the proposed approach once new processes are contemplated in the hyper-model.

\begin{table}[]
\centering
\caption{Ranking of deep drawing processes by shortest distance from process with T=1.5, IS=100 and S=130, using 1) Euclidean Distance and 2) Euclidean Distance with normalized process conditions.}
\label{ranking_processes}
\begin{tabular}{l|lll|lll}
\hline
      & \multicolumn{3}{c}{Euclidean Distance} & \multicolumn{3}{c}{\begin{tabular}[c]{@{}c@{}}Euclidean Distance using \\ Normalized Task Descriptions\end{tabular}} \\
\hline
Order & Thickness & Initial Stress & Saturation & Thickness                            & Initial Stress                           & Saturation                           \\
\hline
1     & 1.5      & 100            & 165        & 1.5                                 & 100                                      & 165                                  \\
2     & 2        & 100            & 165        & 1.5                                 & 100                                      & 200                                  \\
3     & 1.5      & 100            & 200        & 1.5                                 & 175                                      & 227.5                                \\
4     & 2        & 100            & 200        & 1.5                                 & 175                                      & 288.75                               \\
5     & 1.5      & 175            & 227.5      & 1.5                                 & 175                                      & 355                                  \\
6     & 2        & 175            & 227.5      & 1.5                                 & 250                                      & 325                                  \\
7     & 1.5      & 175            & 288.75     & 1.5                                 & 250                                      & 412.5                                \\
8     & 2        & 175            & 288.75     & 1.5                                 & 250                                      & 500                                  \\
9     & 1.5      & 175            & 355        & 2                                   & 100                                      & 165                                  \\
10    & 2        & 175            & 355        & 2                                   & 100                                      & 200                                  \\
11    & 1.5      & 250            & 325        & 2                                   & 175                                      & 227.5                                \\
12    & 2        & 250            & 325        & 2                                   & 175                                      & 288.75                               \\
13    & 1.5      & 250            & 412.5      & 2                                   & 175                                      & 355                                  \\
14    & 2        & 250            & 412.5      & 2                                   & 250                                      & 325                                  \\
15    & 1.5      & 250            & 500        & 2                                   & 250                                      & 412.5                                \\
16    & 2        & 250            & 500        & 2                                   & 250                                      & 500                                 \\
\hline
\end{tabular}
\end{table}

First, this distance between processes is related with each task description. This way, if we assume the task descriptions take the form of a 3-d vector space due to Thickness (T), Initial Stress (IS) and Saturation (S), and we can define, e.g. T=1.5, IS=100 and S=130 as $\vec{\theta}$ = (1.5,100,130). Based on this vector representation, two different distance functions are used: 1) Euclidean Distance (EUC); 2) Euclidean Distance using normalized task descriptions (NORMEUC). The main difference between those is that normalizing between 0 and 1 eliminates the conditions' degree of magnitude. This leads to two different scenarios where the process models incrementally added to the hyper-model change and the impact of different sorting strategies can be assessed. As an example, Table \ref{ranking_processes} presents the ranking of deep drawing processes according to the two approaches for distance calculation, considering the target process of T=1.5, IS=100 and S=130.

In the next sub-section an assessment of LIN, POL2 and POL3 for the two types of distance measure will be made, where further on the same assessment will be made but with LASSO2, LASSO3, RIDGE2 and RIDGE3.

\subsection{Linear, Degree 2 Polynomial and Degree 3 Polynomial Regression}

One of the main aspects to be assessed is the performance of the hyper-model in predicting the deformable parameters for the SSM in the HPM algorithm. This will allow to compare how close the generated shapes are from its ground truth. Thus, first, each of the 17 process model shapes will be generated according to an increasing number of different process models. This increasing number will start with the 4 closest source models according to the distance measure to be used, and the MSE will be calculated for that case. Then, the 5 closest source models will be used, and again, the MSE calculated. This process goes until all the remaining 16 process model shapes are used in the hyper-model for shape generation. For one modeling technique, e.g. LIN, 13 tests (from 4 shapes to 16) will be made per process model, resulting in a total of $13 tests \times 17 processes = 221 tests$ in order to assess all the process models. Finally, 7 different modeling techniques were used, so 221 x 7 = 1547 tests were performed.

\begin{figure}[h]
\vskip 0.2in
\begin{center}
\centerline{\includegraphics[width=5.5in]{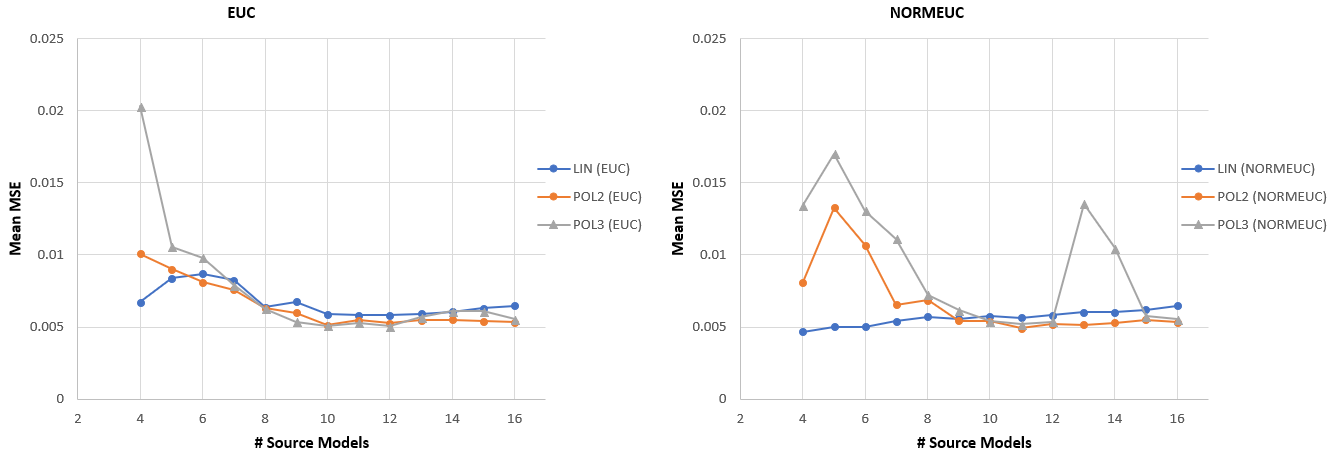}}
\caption{Mean values of MSE for all the processes per number of source models using LIN, POL2 and POL3 techniques. On the left, the Mean MSE for Euclidean distance (EUC) and on the right for the Euclidean distance with normalized process conditions (NORMEUC).}
\label{linear_pol2_pol3_all}
\end{center}
\vskip -0.2in
\end{figure}

The first analysis made was to assess how the error (MSE) between the generated shape and the ground truth evolves as soon as more process models are added to the hyper-model in HPM. Therefore, the mean value of all MSE using a fixed number of source models (e.g. 4) was calculated for all target process models. Figure \ref{linear_pol2_pol3_all} presents all mean values for each number of closest source models, from 4 to 16, for the LIN, POL2 and POL3 techniques using both distance measures EUC and NORMEUC. As can be seen, the first major difference between NORMEUC and EUC approaches is on LIN technique, where different error evolution behaviors exist in regard to the number of source models. In this case, the mean MSE for NORMEUC using 4 to 10 source models is lower when compared with the EUC approach. Once the number of source models increase, the performance of the both approaches tend to be the same, as expected. However, despite the low MSE for a low amount of source models in the NORMEUC, the MSE increases as soon as new source models are added. Additionally, both POL2 and POL3 suffer from overfitting for a lower amount of source models used, leading to high peaks of MSE in both EUC and NORMEUC. For the POL3 approach using NORMEUC, also overfitting effects can be seen for a higher amount of source models. Despite not being significantly lower, the POL2 and POL3 techniques can perform better than LIN in both EUC and NORMEUC for a high number of source models, where it seems to neither increase or decrease. This is contrary to the LIN approach where the MSE is slowly increasing. This slow increase might be due to the fact that a linear model is underfitting the data, revealing that the model complexity is not enough to mimic the system dynamics. Finally, for this first analysis, it can be seen that MSE varies according to the number of source models used, and also that different performances are observed when using different types of statistical techniques for modeling.

\begin{figure}[h!]
\vskip 0.2in
\begin{center}
\centerline{\includegraphics[width=5in]{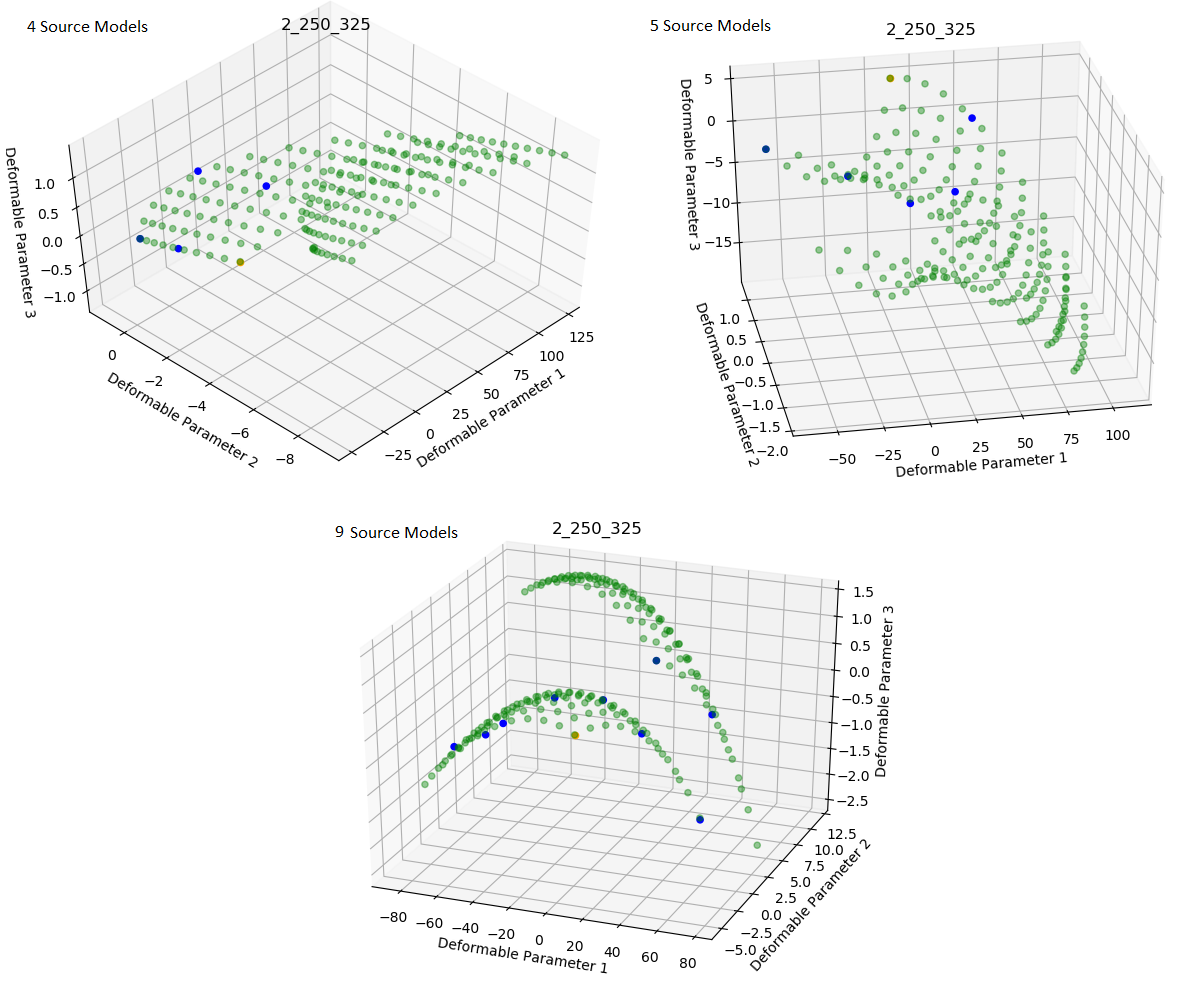}}
\caption{Scatter Plot for the deformable parmaters in the SSM using NORMEUC strategy. The blue points are the source models, the green points are all the predicted values in the range of process conditions (Thickness = [1.5, 2], Initial Stress = [100, 250], Saturation = [130, 500]) and orange point is the predicted value for the target task description (process 15) that will be used to generate the corresponding shape.}
\label{pol2_15_normeuc_allp}
\end{center}
\vskip -0.2in
\end{figure}

In order to confirm if the observed peaks are from overfitting effect, a more thorough analysis should be made. For this case, the target process with conditions Thickness = 2, Initial Stress = 250 and Saturation = 325 was chosen with POL2 used for the hyper-model using the NORMEUC strategy. This process was chosen because it is one of the processes with high MSE for a low amount of source models. The MSE per target model using the range of source models from 4 to 16 for all techniques is not depicted due to space limitation. Figure \ref{pol2_15_normeuc_allp} presents three different plots of deformable parameters, each using different amounts of source models, namely 4 source models (top left plot), 5 source models (top right plot) and 9 source models (bottom plot), where the blue dots are the training datapoints and green ones are the prediction for a wider range of process conditions, assessing the model generalization, and finally the orange ones are the predicted value for the target task description made by the hyper-model. These values of source models were chosen because represent the state of the hyper-model before overfitting, when it overfits and after overfitting, correspondingly. As can be seen, the shape of the predicted values for 4 and 9 source models is simple and smooth, as the presented data points. The observed predicted shape for 5 source models is irregular and it predicts the target process condition far from the optimal value due to this fitting of data using an irregular shape.

\begin{figure}[h!]
\vskip 0.2in
\begin{center}
\centerline{\includegraphics[width=5.5in]{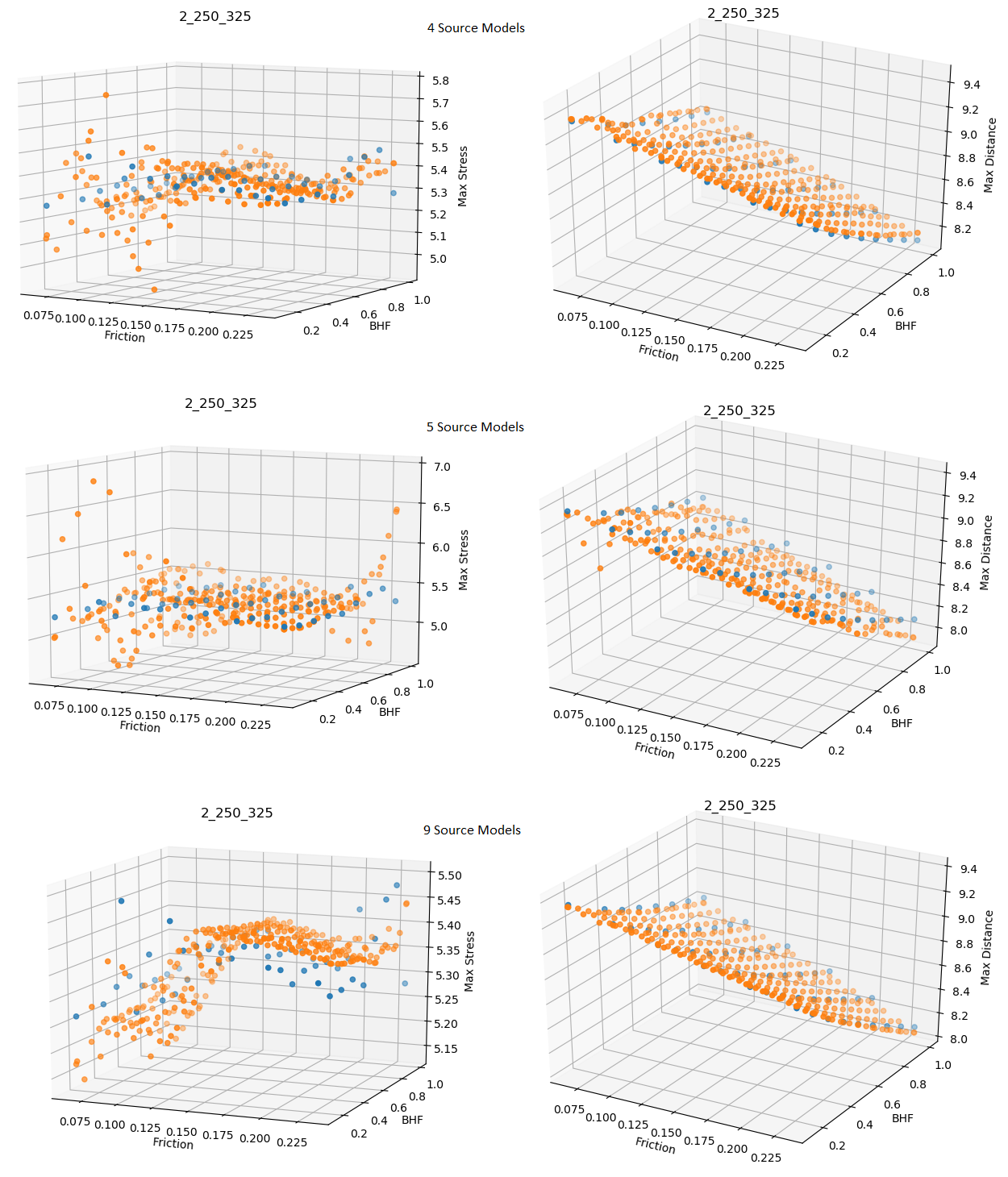}}
\caption{6 Scatter Plots of Maximum Stress and Distance using the for Degreee 2 Polynomial with NORMEUC strategy for shape generation of process 15. On the left there are the Maximum Stress plots with 4, 5 and 9 source models from top to bottom. On the right, the same organization is used but for process quality Distance. The blue points represent the group truth of Maximum Stress and Distance, and the orange points represent the generated shapes from the Hyper-Process Model.}
\label{pol2_15_normeuc_allp_all}
\end{center}
\vskip -0.2in
\end{figure}

Additionally, and in order to better understand the impact of overffiting in the final generated shape, Figure \ref{pol2_15_normeuc_allp_all} presents the Maximum Stress and Distance for all generated shapes using 4, 5 and 9 source models. The blue points represent the ground truth of Maximum Stress and Distance, and the orange points represent the generated shapes from the HPM. It can be seen that the generated shape for 5 source models is far more disperse with high values in its periphery. This difference can be greatly noticed if we look at the value range using 5 source models, where the Maximum Stress varies between 4.5 and 7, and using 4 source models with a shorter range between 4.9 and 5.8. This range is even shorter with 9 source models values between 5.1 and 5.5. As for the process quality Distance, it can be seen that for 5 source models the prediction surface is quite irregular and bumpy compared with the smooth surface from the ground truth. For 4 data points this prediction is better with a more regular surface and for 9 data points it is near perfect fit.

After this, one should now explore why do models overfit in certain situations, and in others don't. This is tightly related with the two used strategies of EUC and NORMEUC. As for the NORMEUC, all the processes with the same Thickness are presented first to train the hyper-model, as seen in Table \ref{ranking_processes}. This means that more heterogeneity can be grasped from Initial Stress and Saturation, which is beneficial only for a specific Thickness in the target process condition. In the EUC case, the same Initial Stress and Saturation is presented first, meaning that initially a high heterogeneity of Thickness is presented to the hyper-model. Assuming that the deep drawing deformable parameters are grouped by Thickness, as observed in Figure \ref{deformable_parameters}, the EUC can be seen as a strategy that initially looks for the same examples in different groups to get an holistic picture of the whole deep drawing. The NORMEUC first looks to all possible examples in the same group to rapidly grasp a local perspective of the process for that specific Thickness. This justifies why MSE for the EUC strategy starts with high values and quickly decreases as soon as new source models are included, but no ovefitting effect occurs due to this more complete understanding of the deep drawing process, and data is more complex. Contrary to this, since in NORMEUC only process models with the same Thickness are included first, a more local perspective and less holistic overview is obtained. This implies that for low amount of source models the MSE is low for LIN because data is far simpler than EUC, but due to low amount of data, when using more complex techniques like POL2 and POL3, overfitting might occur. Nevertheless, since a local perspective is first built for NORMEUC, there's a chance that overfitting can also occur for high number of source models. This can be seen in Figure \ref{linear_pol2_pol3_all} for POL3 using NORMEUC, where it peaks with high MSE at 13 and 14 source models.

In the next sub-section the regularized regression techniques of LASSO and Ridge will be assessed in the same conditions, where it is expected to significantly reduce the overfitting effect from the presented results so far.



\subsection{LASSO Degree 2 Polynomial, LASSO Degree 3 Polynomial, Ridge Degree 2 Polynomial, Ridge Degree 3 Polynomial}


Based on this, the present sub-section is dedicated to the use of LASSO and Ridge regression techniques to assess if the overfitting effects can be attenuated and a better performance can be achieved. LASSO uses a L1-norm regularization term, as for the Ridge regression a L2-norm is used. In these techniques the $\lambda$ parameter controls the amount of shrinkage applied. For high values of $\lambda$, the shrinkage is also higher, as if the $\lambda$ decreases to 0, less penalization is applied. The main difference between these techniques is that Ridge tends to minimize the value of model parameters to be near zero, while LASSO tends to shrink the parameters to zero, discarding irrelevant parameters in the regression problem \cite{friedman2001elements}.


Therefore, 4 different approaches were applied using NORMEUC and EUC strategies. These are LASSO for degree 2 and 3 polynomial, as well as Ridge with the same polynomial degrees. Several values of $\lambda$ were tested in order to find the best hyper-parameter that minimized the MSE for an increasing number of source models used to generate the target shapes. The strategy used in order to find the best $\lambda$ starts with the high penalization value of 0.1 and progressively decreasing this value by a factor of 10 until the overall MSE starts to increase.

\begin{figure}[h!]
\vskip 0.2in
\begin{center}
\centerline{\includegraphics[width=5.5in]{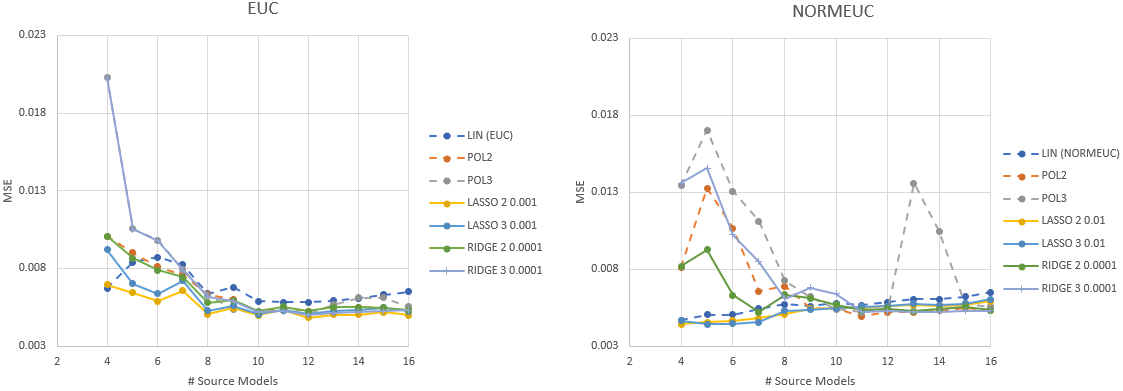}}
\caption{Plot of all techniques tested for each of the strategies used to measure distance among process models: EUC on the left and NORMEUC on right. All the penalization $\lambda$ used for LASSO2, LASSO3, Ridge2 and Ridge3 minimized the overall MSE.}
\label{euc_normeuc_complete}
\end{center}
\vskip -0.2in
\end{figure}

For this analysis, first we will start with the EUC strategy and assessing all the proposed techniques, and secondly, the same assessment will be performed but for NORMEUC. Hence, Figure \ref{euc_normeuc_complete} depicts the results of the analysis performed, where on the left the results for the EUC strategy are depicted and NORMEUC on the right side. In this case, the already shown graphs for LIN, POL2 and POL3 are presented as dashed lines for ease of reading.

This way, starting with EUC, one major behavior draws the attention. As expected, despite the performance variation for a small amount of models, as soon as new models start being added to the hyper-model for training, the performance of all techniques tend to be the same. Moreover, the benefits of LASSO regression over the three techniques LIN, POL2 and POL3 can be clearly seen. Due to regularization the overfitting effect can be mitigated using both LASSO2 and LASSO3, where LASSO2 is the best option for small amount of models. As already discussed, for a small amount of source models, LIN fails to grasp a great majority of the details due to low model complexity leading to underfitting, where in POL2 and POL3 overfitting is observed due to low amount of datapoints. Hence, it is seen that LASSO can handle such situations. For both techniques, the best value for the penalization $\lambda$ is 0.001. As for the Ridge regression with degree 2 and 3 polynomials, no significant benefits were observed when compared with POL2 and POL3, correspondingly, where the best value in both techniques for penalization $\lambda$ is 0.0001. Only a small decrease in the MSE between 13 and 15 source models when comparing to POL3 (grey dashed line) was achieved. Additionally, for LIN regression a small increase of MSE with a high amount of models is observed, mainly due to small model complexity that fails to grasp some details of system dynamics, leading to underfitting. Hence, in all techniques used, this underfitting effect is diminished due to the higher complexities from the techniques used. In sum, the main conclusion for the EUC strategy is related with the benefits of using regularization for a small amount of models, in which overfitting is avoided, and at the same time avoiding underfitting for high amount of source models. This way, for this scenario, regularization brings the best out of low and higher complexity models. For the proposed scenario, the best technique that can be used is LASSO2.

Regarding strategy NORMEUC, again the same overall behavior is observed, where for a high amount of models the performance of all the techniques tend to be the same. As for LASSO2 and LASSO3 the best value for penalization is 0.01 and a significant improvement over POL2 and POL3, and even LIN, is observed. A low value of MSE for a small amount of source models to train the hyper-model is achieved demonstrating once again the great benefits of regularization in case of overfitting. As for a higher number of source models, these techniques also demonstrate a better performance than LIN, POL2 and POL3, excepting for some particular values of source models. The techniques of Ridge2 and Ridge3 used the best $\lambda$ value of 0.0001 and improve the performance of the system when compared with POL2 and POL3, correspondingly, although not as good as the LASSO2 and LASSO3. Nevertheless, the effects of overfitting are reduced, and in the case of high MSE peak in POL3 at 13 and 14 source models, the overfitting is completely eliminated. However, one can see an overall strange behavior that does not happen in EUC, which is an increase in MSE from 7 and 8 source models used for training. As already explained when solely using LIN, POL2 and POL3, this behavior occurs because NORMEUC first presents a group of source models with the same value of Thickness as the target model (one cluster of data), and from 7 (for target Thickness of 2) and 8 (for target Thickness of 1.5) the other group of source models with a different Thickness is presented.


\begin{figure}[h!]
\vskip 0.2in
\begin{center}
\centerline{\includegraphics[width=5in]{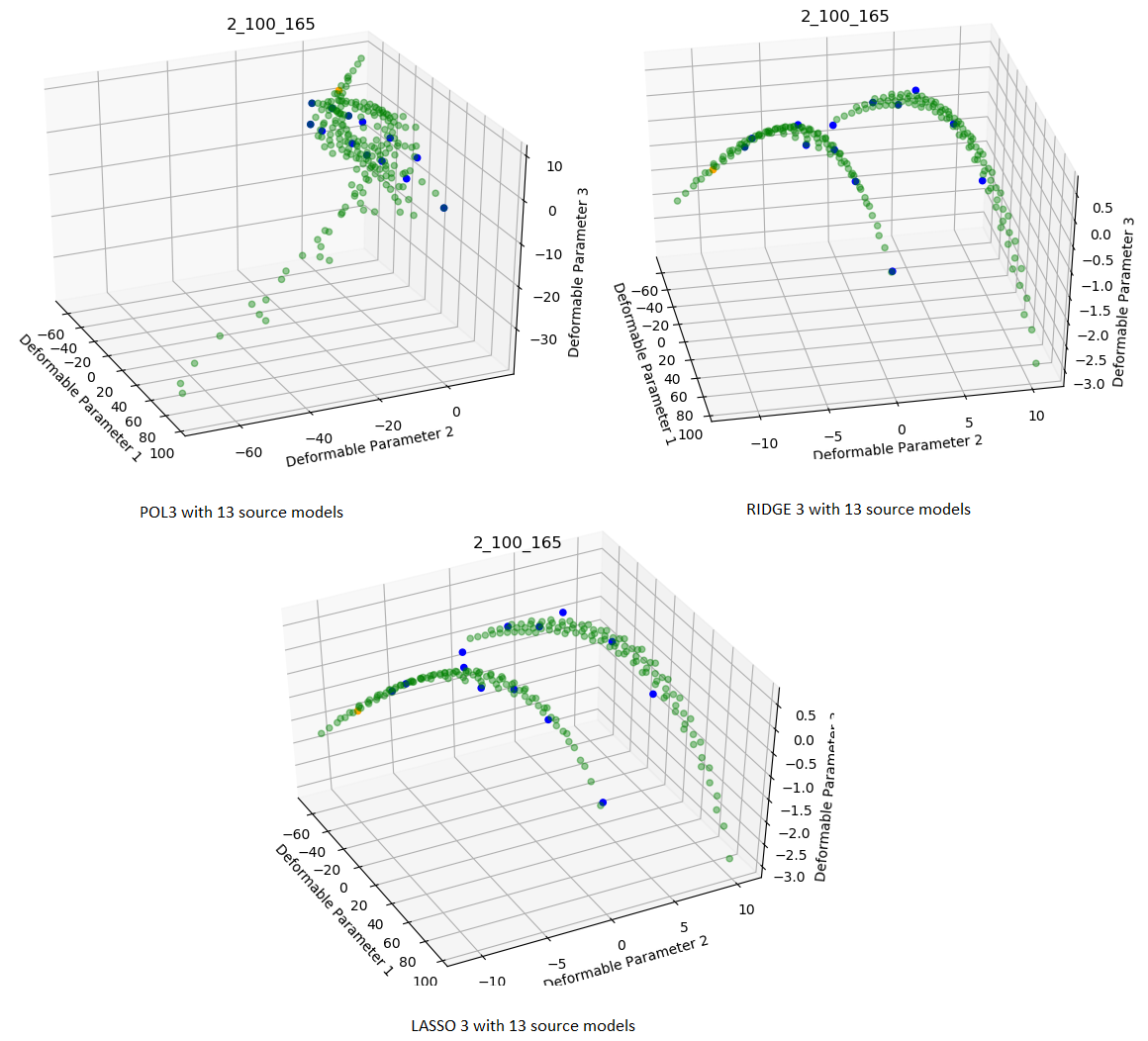}}
\caption{Deformable parameter space for both POL3 (left), RIDGE 3 (right) and LASSO 3 (bottom) using NORMEUC strategy.}
\label{all_normeuc_10_13p}
\end{center}
\vskip -0.2in
\end{figure}


In order to prove that LASSO3 and Ridge3 can alleviate the overfitting effect by penalizing high values of the hyper-model parameters, a visual inspection should be made in the deformable parameter space. For this case, the target process with task descriptors T = 2, IS = 100 and S = 165 using 13 source models for NORMEUC strategy and POL3 for training the hyper-model were used. These conditions were chosen because led to hyper-model overfitting where high peaks of MSE occur. Figure \ref{all_normeuc_10_13p} depicts on the left the deformable parameter values using POL3 technique, on the right the same plot is shown but for RIDGE 3 and LASSO 3 on the bottom. Hence, each plot presents the corresponding deformable parameters used to train the hyper-model (blue dots), together with the predicted values for all the range of task descriptions (green) and the prediction of deformable parameter for the target task description (orange). As observed, the produced deformable parameter surface for POL3 is very complex and can be clearly seen that is not a good fit and generalization. On the contrary, for the RIDGE3 and LASSO3, the produced surface is very smooth, simple and it represents and good generalization, clearly grasping the shape of data. 



\section{Discussion and Main Conclusions}

From the results presented previously, we would like to highlight two different aspects that significantly influence the system performance. The first aspect is related with which source models should be used to build the hyper-model and generate a new target model. As it was observable from the two strategies that incrementally define which source models should be used to train the hyper-model, namely EUC and NORMEUC, the selected models can benefit or hamper the final results. As expected, as the number of models increase the tendency is for both strategies to converge for the same performance, since almost the same models are is both training sets. Although, this is not true by using a smaller amount of models. In this case, we observe that in order to maximize the performance of the system, the most suitable strategy for source model selection will benefit the overall system performance. In the present work, only a distance based approach using the process conditions was used, but others might be more adequate, depending on the application case. Since one of the advantages of the HPM is to learn a set of unsupervised features based on data shape and modes of variation, the distance among the feature vectors would be a good similarity indication. As can be seen from Figure \ref{deformable_parameters}, there are some correlations and data clusters between the learned deformable parameters. One can take advantage on these to estimate a set of source models to use in the HPM algorithm.


The second aspect that greatly influences the system performance is the possibility of, what we call, \textit{domain drift}. This drift can occur due to the two-stage learning approach of the HPM. In the first stage, source models should be trained with the available datasets and in the second one, a hyper-model should be trained to map task descriptions into deformable parameters. From the proposed algorithm, it is not guaranteed that both stages are optimal, leading to domain drift. This effect was one of the limitations from DAP (one of the firsts ZSL algorithms) that was tackled by the ALE algorithm proposed by \citet{akata2013label} by performing a joint learning of input and output embeddings. As referred in the Section \ref{intro_rw}, this joint learning is performed by using bi-linear compatibility functions where the compatibility among these embeddings is maximized. This domain drift effect was observed when the trained hyper-model overfitted the data when using both degree 2 and 3 polynomials. This difference in performances from the first to the second stage was addressed mainly with LASSO, where the regularized attenuated the overfitting effect, minimizing the domain drift. 

Meanwhile, if one compares the state of the art methods of ZSL for regression, such as the hyper-model or TaDeLL, with the HPM algorithm, one can easily conclude that those are much more susceptible to domain drift than HPM. Since in TaDeLL or hyper-model alone only one technique can be used to model all the source tasks to keep the parameters in the same feature space, the probability of getting a sub-optimal model in the first stage is higher than in HPM. As different tasks can have different properties, being some more complex than others, the same technique can be either a good or bad candidate for training, according to these properties. By taking advantage on HPM being independent from the techniques used to train models in the first stage, at principle optimal or near-optimal models can be achieved. This eases the problem of domain drift by only addressing overfitting effects on the second stage, which can be tackled with regularized techniques, as previously demonstrated.

Online citations: \cite{deepdrawing}


\bibliographystyle{ACM-Reference-Format}
\bibliography{sample-bibliography}

\end{document}